\documentclass[10pt,journal,compsoc]{IEEEtran}
\usepackage[pdftex]{graphicx}
\usepackage{booktabs, makecell, tabularx}
\usepackage{rotating}
\usepackage[export]{adjustbox}
\usepackage{gensymb}
\usepackage{ragged2e}

\ifCLASSOPTIONcompsoc
  \usepackage[nocompress]{cite}
\else
  \usepackage{cite}
\fi
\ifCLASSINFOpdf
  \usepackage[pdftex]{graphicx}
  \graphicspath{{../pdf/}{../jpeg/}{./figures}}
\else
\fi
\usepackage{xspace}
\usepackage{pgfplots}
\usepackage{xcolor}
\usepackage{tikz}

\usepackage{amsmath}
\usepackage{amssymb}
\usepackage{array}

\ifCLASSOPTIONcompsoc
  \usepackage[caption=false,font=footnotesize,labelfont=sf,textfont=sf]{subfig}
\else
  \usepackage[caption=false,font=footnotesize]{subfig}
\fi

\ifCLASSOPTIONcaptionsoff
  \usepackage[nomarkers]{endfloat}
 \let\MYoriglatexcaption\caption
 \renewcommand{\caption}[2][\relax]{\MYoriglatexcaption[#2]{#2}}
\fi
\usepackage{url}
\usepackage{etex}

\renewcommand{\baselinestretch}{.985}

\begin{document}
\bstctlcite{IEEEexample:BSTcontrol}
%

\title{Gliding vertex on the horizontal bounding box for multi-oriented object detection}

%
%
%
%

\author{Yongchao Xu,
        Mingtao Fu,
        Qimeng Wang,
        Yukang Wang,
        Kai Chen, \\
        Gui-Song Xia~\IEEEmembership{Senior Member,~IEEE},
        Xiang Bai~\IEEEmembership{Senior Member,~IEEE}
\IEEEcompsocitemizethanks{\IEEEcompsocthanksitem Y. Xu, M. Fu, Q. Wang, Y. Wang, and X. Bai are with the School
of Electronic Information and Communications, Huazhong University of
Science and Technology (HUST), Wuhan, 430074, China. \protect\\
E-mail: \{yongchaoxu, mingtaofu, qimengwang, wangyk, xbai\}@hust.edu.cn.
\IEEEcompsocthanksitem G. S. Xia is with LIEMARS, Wuhan University. \protect\\
E-mail: guisong.xia@whu.edu.cn.
\IEEEcompsocthanksitem K. Chen is with Shanghai Jiaotong University; Onyou Inc. \protect\\
E-mail: kchen@sjtu.edu.cn.
}
}

%
%

\markboth{Journal of \LaTeX\ Class Files,~Vol.~XX, No.~XX, September~2019}%
{Shell \MakeLowercase{\textit{et al.}}: Bare Demo of IEEEtran.cls for Computer Society Journals}
%



\newcommand{\ie}{\textit{i.e.}\xspace}
\newcommand{\eg}{\textit{e.g.}\xspace}
\newcommand{\etal}{\textit{et al.}\xspace}
\newcommand{\wrt}{\textit{w.r.t.}\xspace}
\newcommand{\etc}{\textit{etc.}\xspace}
\newcommand{\resp}{\textit{resp.}\xspace}

\newcommand{\fixedvskip}{-3mm}

\newcommand{\fixedvskiptab}{-2mm}

\IEEEtitleabstractindextext{%
\begin{abstract}
  \justifying
Object detection has recently experienced substantial progress. Yet,
the widely adopted horizontal bounding box representation is not
appropriate for ubiquitous oriented objects such as objects in aerial
images and scene texts. In this paper, we propose a simple yet
effective framework to detect multi-oriented objects. Instead of
directly regressing the four vertices, we glide the vertex of the
horizontal bounding box on each corresponding side to accurately
describe a multi-oriented object. Specifically, We regress four length
ratios characterizing the relative gliding offset on each
corresponding side. This may facilitate the offset learning and avoid
the confusion issue of sequential label points for oriented objects.
To further remedy the confusion issue for nearly horizontal objects,
we also introduce an obliquity factor based on area ratio between the
object and its horizontal bounding box, guiding the selection of
horizontal or oriented detection for each object. We add these five
extra target variables to the regression head of faster R-CNN, which
requires ignorable extra computation time. Extensive experimental
results demonstrate that without bells and whistles, the proposed
method achieves superior performances on multiple multi-oriented
object detection benchmarks including object detection in aerial
images, scene text detection, pedestrian detection in fisheye
images.

\end{abstract}

}

\maketitle

\IEEEdisplaynontitleabstractindextext

%
\IEEEpeerreviewmaketitle

\IEEEraisesectionheading{\section{Introduction}\label{sec:introduction}}

\IEEEPARstart{O}{bject} detection has achieved a considerable progress
thanks to convolutional neural networks (CNNs). The state-of-the-art
methods~\cite{ren2017faster, redmon2018yolov3, lin2017feature} usually
aim to detect objects via regressing horizontal bounding boxes. Yet
multi-oriented objects are ubiquitous in many scenarios. Examples are
objects in aerial images and scene texts. Horizontal bounding box does
not provide accurate orientation and scale information, which poses
problem in real applications such as object change detection in aerial
images and recognition of sequential characters for multi-oriented
scene texts.

Recent advances in multi-oriented object detection are mainly driven
by adaption of classical object detection methods using rotated
bounding boxes~\cite{zhou2017east, ding2018transformer} or
quadrangles~\cite{liao2018textboxes++, he2018direct, zhang2019look} to
represent multi-oriented objects. Though these existing adaptions of
horizontal object detection methods to multi-oriented object detection
have achieved promising results, they still face some limitations. For
detection using rotated bounding boxes, the accuracy of angle
prediction is critical. A minor angle deviation leads to important IoU
drop, resulting in inaccurate object detection. This problem is more
prominent for detecting long oriented objects such as bridges and
harbors in aerial images and Chinese text lines in scene images. The
methods based on quadrangle regression usually have ambiguity in
defining the ground-truth order of four vertices, yielding unexpected
detection results for objects of some orientations.

Some other methods~\cite{shi2017seglink, lyu2018multi, liu2018mcn}
alternatively detect horizontal object parts followed by a grouping
process. Yet, such grouping process step is usually heuristic and
time-consuming. Describing an oriented object as its segmentation
mask~\cite{zhang2016multi} is another alternative solution. However,
this often results in split and/or merged components, requiring a
heavy and time-consuming post-processing.

\begin{figure}
\centering
\includegraphics[width=\linewidth]{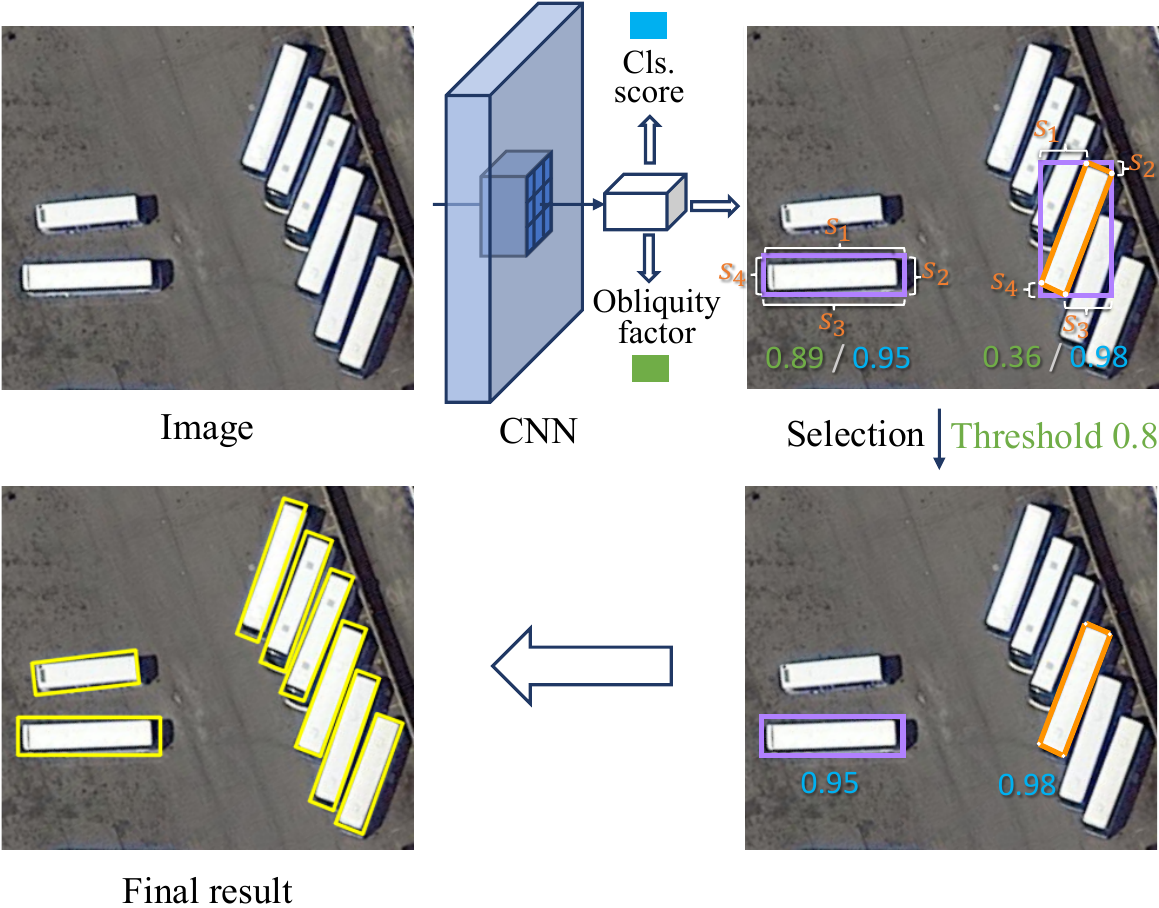}
\vspace{-6mm}
\caption{Pipeline of the proposed method. An image is fed into a CNN,
  which outputs a classification score (blue value), a horizontal
  bounding box, four length ratios between each segment $s_i$ and
  corresponding side, and an obliquity factor (green value) for each
  detection. Based on obliquity factor, we select horizontal box (in
  purple) or oriented detection (in orange) as the final result. Best
  viewed in electronic version.}
\label{fig:pipeline}
\end{figure}

In this paper, we propose a simple yet effective framework to deal
with multi-oriented object detection. Specifically, we propose to
glide each vertex of the horizontal bounding box on the corresponding
side to accurately describe a multi-oriented object. This results in a
novel representation by adding four gliding offset variables to
classical horizontal bounding box representation. Put it simply, we
regress four length ratios that characterize the relative gliding
offset (see Fig.~\ref{fig:pipeline}) on each side of horizontal
bounding box. Such representation may be less sensitive to offset
prediction error than angle prediction error in rotated bounding box
representation. By limiting the offset on the corresponding side of
horizontal bounding box, we may facilitate offset learning and also
avoid the confusion for sequential label points in directly regressing
the four vertices of oriented objects.
To further get rid of confusion issue for nearly horizontal objects,
we also introduce an obliquity factor based on area ratio between
the multi-oriented object and its horizontal bounding box. As depicted
in Fig.~\ref{fig:pipeline}, this obliquity factor guides us to select
the horizontal detection for nearly horizontal objects and oriented
detection for oriented objects. It is noteworthy that the proposed
method only introduces five additional target variables, requiring
ignorable extra computation time.

In summary, the main contribution of this paper are three folds: 1) We
introduce a simple yet effective representation for oriented objects,
which is rather robust to offset prediction error and does not have
the confusion issue. 2) We propose an obliquity factor that
effectively guides the selection of horizontal detection for nearly
horizontal objects and oriented detection for others, remedying the
confusion issue for nearly horizontal objects. 3) Without bells and
whistles (\eg, cascade refinement or attention mechanism), the
proposed method outperforms some state-of-the-art methods on multiple
multi-oriented object detection benchmarks.


\section{Related Work}
\label{sec:related}

\subsection{Deep general object detection}\label{subsec:general_object_detection}

Object detection aims to detect general objects in images with
horizontal bounding boxes. \textcolor{black}{Recent mainstream
  CNN-based methods} can be roughly summarized into top-down and
bottom-up methods. Top-down methods directly detect entire
objects. They can be further categorized into two classes: two-stage
and single-stage methods.  R-CNN and its
variances~\cite{girshick2014rich, girshick2015fast,
  ren2017faster,dai2016rfcn,lin2017feature} are representative
two-stage methods. They first generate object proposals and
then use the features of these proposals to predict object categories
and refine the bounding boxes.
YOLO and its variances~\cite{redmon2016you, redmon2017yolo9000,
  redmon2018yolov3}, SSD~\cite{liu2016ssd}, and
RetinaNet~\cite{lin2017focal} are representative single-stage
methods. They predict bounding boxes directly from deep feature maps
instead of region proposals. Bottom-up methods rise recently by
predicting object parts followed by a grouping process.
CornerNet~\cite{law2018cornernet}, ExtremeNet~\cite{zhou2019bottom},
and CenterNet~\cite{duan2019centernet} are recently proposed in
succession. They attempt to predict some keypoints of objects such as
corners or extreme points, which are then grouped into bounding
boxes. Center points are also used by~\cite{zhou2019bottom,
  duan2019centernet} as supplemental information for grouping.

\subsection{Multi-oriented object detection}
\label{subsec:multi_oriented_object}

\noindent\textbf{Object detection in aerial images}
is chanllenging because of huge
scale variations and arbitrary orientations.
Extensive studies have been devoted to this task. The baselines on the
popular dataset DOTA~\cite{xia2018dota} replace horizontal box
regression of faster R-CNN with regression of four vertices of
quadrangle representation. Many methods resort to rotated bounding box
representation. Rotated RPN is exploited in~\cite{liu2017rrpnship,
  zhang2018toward}, which involves more anchors and thus requires more
runtime.
Ding \etal~\cite{ding2018transformer} propose an RoI transformer that
transforms horizontal proposals to rotated ones, on which the rotated
bounding box regression is performed. Azimi
\etal~\cite{azimi2018towards} adopt an image-cascade network to
extract multi-scale features. Yang \etal~\cite{yang2018r2cnn++} employ
multi-dimensional attention to extract robust features, better coping
with complex backgrounds. Zhang \etal~\cite{zhang2019cad} propose to
learn global and local contexts together to enhance the features.

\smallskip
\noindent\textbf{Oriented scene text detection}
is a challenging problem due to arbitrary orientations.
The mainstream CNN-based detectors can be roughly
divided into regression-based and
segmentation-based~\cite{zhang2016multi, xu2019textfield} methods.
We \textcolor{black}{focus on} regression-based methods.
Most methods directly predict entire texts
using rotated bounding box or quadrangle representation. Ma
\etal~\cite{ma2018arbitrary} employ rotated RPN in the framework of
faster R-CNN~\cite{ren2017faster} to generate rotated proposals and
further perform rotated bounding box regression. Liu
\etal~\cite{liu2017match} propose to use quadrangle sliding windows to
match texts with perspective
transformation. TextBoxes++~\cite{liao2018textboxes++} adopts vertex
regression on SSD~\cite{liu2016ssd}. RRD~\cite{liao2018rotation}
further improves TextBoxes++~\cite{liao2018textboxes++} by decoupling
classification and bounding box regression on rotation-invariant and
rotation-sensitive features, respectively, making the regression more
accurate for long texts.  Both EAST~\cite{zhou2017east} and Deep
direct regression~\cite{he2018direct} perform rotated bounding box
regression and/or vertex regression at each location.

\smallskip
\noindent\textbf{Pedestrian detection in fisheye images}
is different from general
pedestrian detection because pedestrians in fisheye images are
often multi-oriented. Seidel \etal~\cite{seidel2018omnidetector}
propose to transform omnidirectional images into perspective ones, on
which the detection is applied. Such transformation introduces extra
computation time. Based on the prior knowledge that objects in fisheye
images are radial, Tamura \etal~\cite{tamura2019omnidirectional}
propose to train a general object detector with rotated images and
then determine the orientations \textcolor{black}{based on the relative
  positions of object centers \textit{w.r.t.} the image center.}

\subsection{Comparison with related works}
\label{subsec:comparison_with_related}
Compared with the related works, the proposed method targets on
general and ubiquitous multi-oriented object detection with a simple
yet effective framework. By gliding the vertex of horizontal bounding
box on each corresponding side and a novel divide-and-conquer
selection scheme for nearly horizontal and oriented objects, the
proposed method may better learn the offset for accurate
multi-oriented object detection and does not suffer from confusion
issue.
Furthermore, the proposed method may be complementary and easily
plugged into many existing methods focusing on enhancing features. To
equip them with the proposed approach, we only need to replace rotated
bounding box or vertex regression by regressing the four length ratios
and obliquity factor in addition to horizontal bounding box. Such
modification requires ignorable extra runtime.

\section{Proposed Method}
\label{sec:method}

\subsection{Overview}
\label{subsec:overview}
CNN-based object detectors perform well on detecting horizontal
objects but struggle on oriented ones, in particular for long and
dense oriented objects. Direct adaption using rotated bounding box
$B_r$ regression tends to produce inaccurate results due to high
sensitivity to angle prediction error. Regressing the four vertices of
quadrangle representation does not suffer from this problem, but also
fails on some cases because of the ambiguity in defining the order of
four ground truth vertices to be regressed. We attempt to solve the
general multi-oriented object detection by introducing a simple
representation for oriented objects and a novel detection scheme that
divides and conquers nearly horizontal and oriented object detection,
respectively. Specifically, we propose to glide the vertex of
horizontal bounding box $B_h$ on each corresponding side to accurately
describe an oriented object. Put it simply,
in addition to $B_h$, we compute four length ratios that characterize
the relative gliding offset on each side of $B_h$. Besides, We also
introduce an obliquity factor based on area ratio between
multi-oriented object and its horizontal bounding box $B_h$. Based on
the estimated obliquity factor, we select the horizontal (\resp
oriented) detection for a nearly horizontal (\resp oriented) object.
This simple yet effective framework only introduces five target
variables compared with classical horizontal object detectors,
requiring ignorable extra computation time.

\subsection{Multi-Oriented object representation}
\label{subsec:proposed_method}

\begin{figure}
	\centering
	\includegraphics[width=0.5\linewidth]{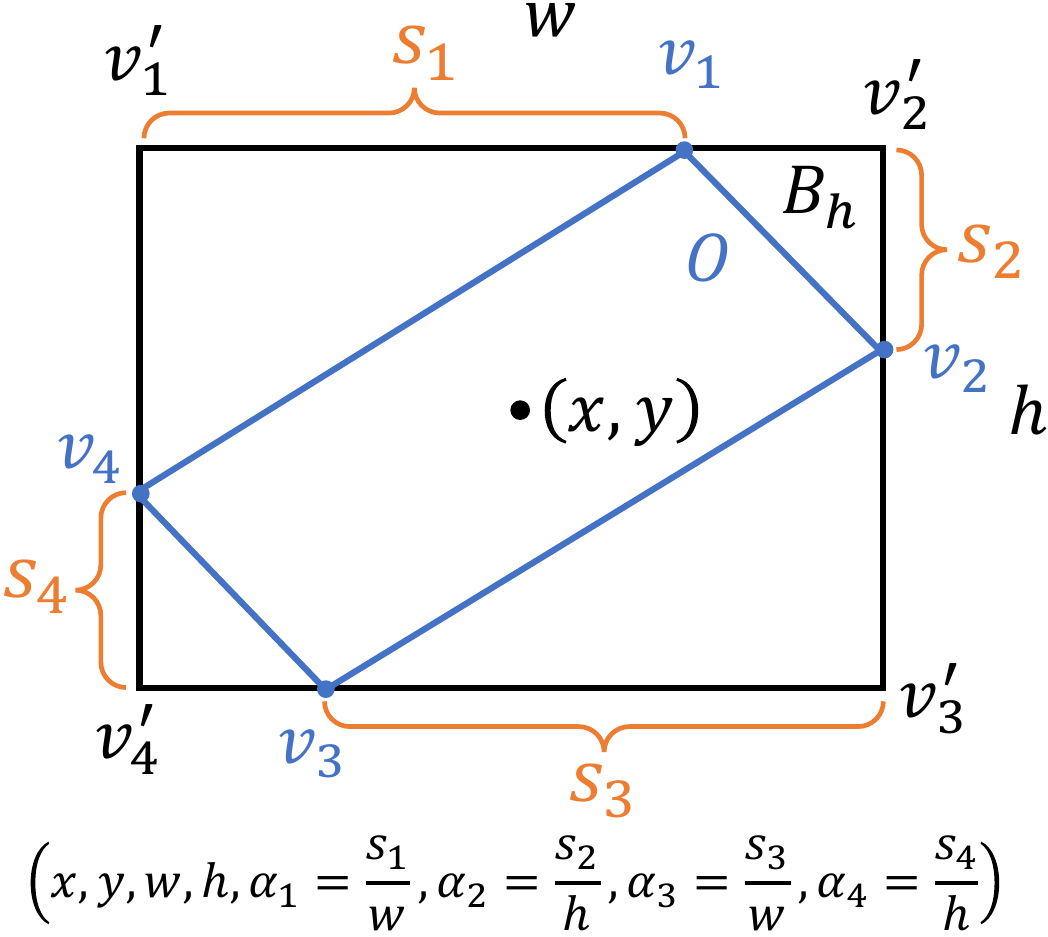}
        \vspace{\fixedvskip}
	\caption{Illustration of proposed representation for an
          oriented object $O$ based on four intersecting points
          $\{v_i\}$ between $O$ and its horizontal bounding box $B_h =
          (v'_1, v'_2, v'_3, v'_4) = (x, y, w, h)$. We adopt $(x, y,
          w, h, \alpha_1, \alpha_2, \alpha_3, \alpha_4)$ to represent
          oriented objects.}
	\label{fig:our}
\end{figure}

The proposed method relies on a simple representation for oriented
objects and an effective selection scheme. An intuitive illustration
of the proposed representation is depicted in Fig.~\ref{fig:our}. For
a given oriented object $O$ (blue box in Fig.~\ref{fig:our}) and its
corresponding horizontal bounding box $B_h$ (black box in
Fig.~\ref{fig:our}), let $v_i, i \in \{1, 2, 3, 4\}$ denote top,
right, bottom, left intersecting point with its horizontal bounding
box $B_h$ denoted by $v'_i, i \in \{1, 2, 3, 4\}$, respectively. The
horizontal bounding box $B_h$ is also usually represented by $(x, y,
w, h)$, where $(x, y)$ is the center, and $w$ and $h$ are the width
and height, respectively. We propose to represent the underlying
oriented object by $(x, y, w, h, \alpha_1, \alpha_2, \alpha_3,
\alpha_4)$. The extra variables $\alpha_i, i \in \{1, 2, 3, 4\}$ are
defined as follows:
\begin{equation}
\begin{split}
\label{eq:alpha}
  \alpha_{\{1,3\}}\;&=\;\|s_{\{1,3\}}\|/w,
  \\ \alpha_{\{2,4\}}\;&=\;\|s_{\{2,4\}}\|/h,
\end{split}
\end{equation}
where $\|s_i\| = \|v_i - v'_i\|$ denotes the distance between $v_i$
and $v'_i$, \ie, the length of segment $s_i = (v_i, v_i')$
representing the gliding offset from $v'_i$ to $v_i$. It is noteworthy
that all $\alpha_i$ is set to 1 for horizontal objects.

In addition to the simple representation in terms of $(x, y, w, h,
\alpha_1, \alpha_2, \alpha_3, \alpha_4)$ for an oriented object $O$,
we also introduce an obliquity factor characterizing the tilt degree
of $O$. This is given by the area ratio $r$ between $O$ and $B_h$:
\begin{equation}
\label{eq:ratio}
r = |O|\;/\;|B_h|,
\end{equation}
where $|\cdot|$ denotes the cardinality. Nearly horizontal objects
have a large obliquity factor $r$ being close to 1, and the obliquity
factor $r$ for extremely slender and oriented objects are close to
0. Therefore, we can select the horizontal or oriented detection as
the final result based on such obliquity factor $r$. Indeed, it is
reasonable to represent nearly horizontal objects with horizontal
bounding boxes. However, oriented detections are required to accurately
describe oriented objects.

\subsection{Network architecture}
\label{subsec:networkarchitecture}

\begin{figure}
\centering
\includegraphics[width=0.85\linewidth]{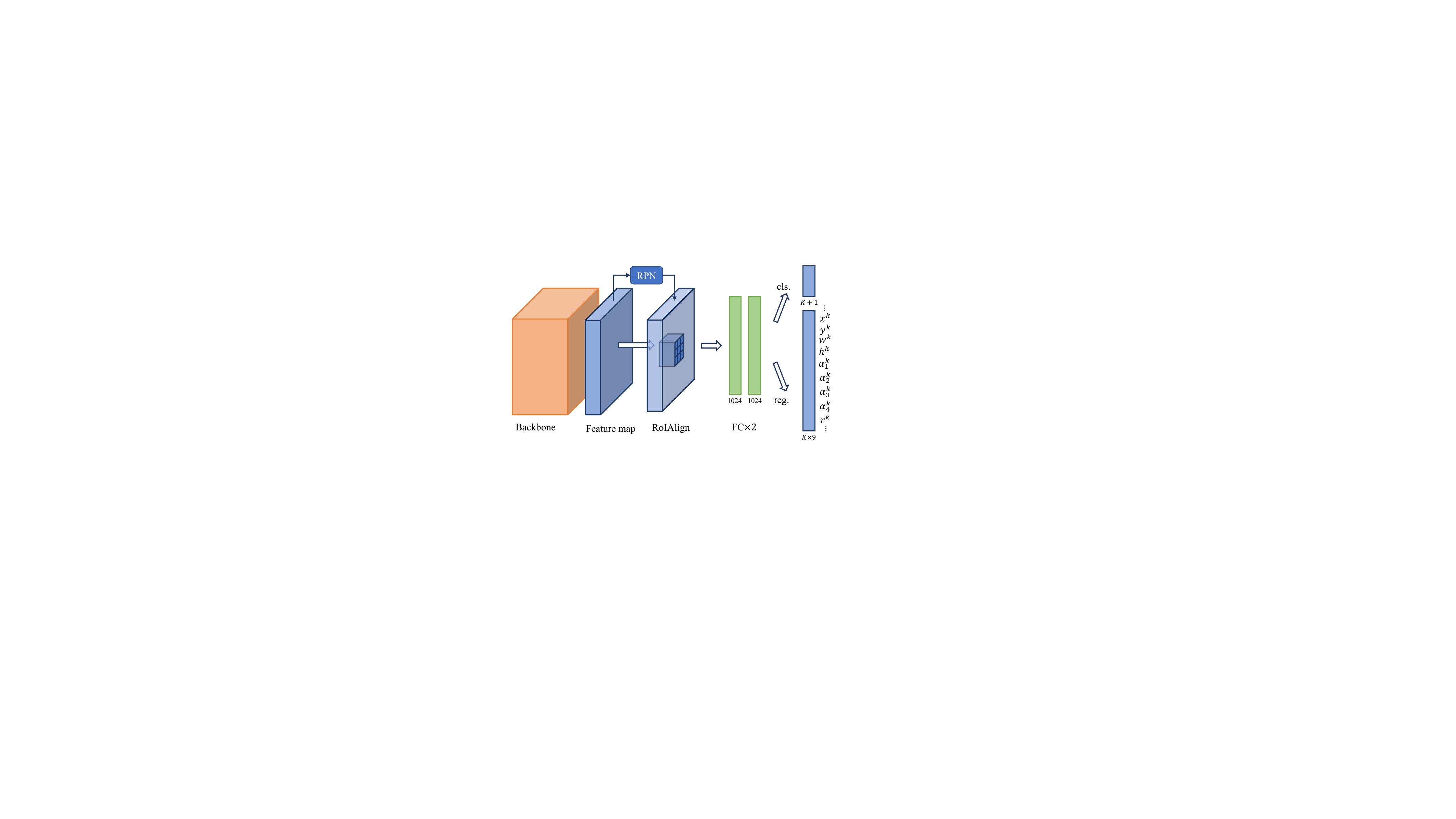}
\vspace{\fixedvskip}
\caption{Network architecture. We simply add five extra target
  variables (normalized to $[0, 1]$ using the sigmoid funciton) to the
  head of faster R-CNN~\cite{ren2017faster}. $K$: number of classes;
  $k$: a certain class.}
\label{fig:architecture}
\end{figure}

The network architecture (see Fig.~\ref{fig:architecture}) is almost
the same as faster R-CNN~\cite{ren2017faster}. We simply add five
extra target variables (normalized to $[0, 1]$ using the sigmoid
funciton) to the head of faster
R-CNN~\cite{ren2017faster}. Specifically, The input image is first fed
into a backbone network to extract deep features and generate bounding
box proposals with RPN~\cite{ren2017faster}. Then the regional
features extracted via RoIAlign~\cite{he2017mask} on proposals are
passed through a modified R-CNN head to generate final results,
including a horizontal bounding box $(x, y, w, h)$, four variables
$(\alpha_1, \alpha_2, \alpha_3, \alpha_4)$ characterizing the oriented
bounding box, and obliquity factor $r$ that indicates whether the
object is nearly horizontal or not.

\subsection{Ground-truth generation}
\label{subsec:gt_gen}

The ground-truth for each object is composed of three components:
classical horizontal bounding box representation $(\tilde{x},
\tilde{y}, \tilde{w}, \tilde{h})$, four extra variables
$(\tilde{\alpha}_1, \tilde{\alpha}_2, \tilde{\alpha}_3,
\tilde{\alpha}_4)$ representing the oriented object, and the obliquity
factor $\tilde{r}$. The horizontal bounding box ground-truth follows
the pioneer work in~\cite{girshick2014rich}, which is relative to the
proposal. The ground-truth for the four extra variables
$(\tilde{\alpha}_1, \tilde{\alpha}_2, \tilde{\alpha}_3,
\tilde{\alpha}_4)$ and obliquity factor $\tilde{r}$ depend only on the
underlying ground-truth object, and are directly calculated by
Eq.~\eqref{eq:alpha} and~\eqref{eq:ratio}, respectively.

\subsection{Training objective}
\label{subsec:trainingobjective}
The proposed method involves loss for RPN stage and R-CNN stage. The
loss of RPN is the same as that in~\cite{ren2017faster}. The loss $L$
for R-CNN head contains a classification loss term $L_{cls}$ and a
regression loss term $L_{reg}$. The R-CNN loss $L$ is given by
\begin{equation}
\label{eq:lossrcnn}
\begin{split}
L\;&=\;\frac{1}{N_{cls}}\sum_{i}{L_{cls}}+\frac{1}{N_{reg}}\sum_{i}{p^{*}_{i} \times L_{reg}},
\end{split}
\end{equation}
where $N_{cls}$ and $N_{reg}$ are the number of total proposals and
positive proposals in a mini-batch fed into the head, respectively,
and $i$ denotes the index of a proposal in a mini-batch. If the $i$-th
proposal is positive, $p^{*}_{i}$ is $1$, otherwise it is $0$. The
regression loss $L_{reg}$ contains three terms for horizontal bounding
box, four length ratios $(\alpha_1, \alpha_2, \alpha_3, \alpha_4)$,
and obliquity factor $r$ regression, respectively. Put it simply, the
regression loss $L_{reg}$ is given by
\begin{equation}
\label{eq:lossreg}
\begin{split}
L_{reg}\;&=\;{\lambda}_{1} \times L_{h} + {\lambda}_{2} \times L_{\alpha} + {\lambda}_{3} \times L_{r}, \\
L_{\alpha}\;&=\;\sum_{i=1}^{4}{{\rm smooth}_{L_1}(\alpha_i-\tilde{\alpha}_i)}, \\
L_{r}\;&=\;{\rm smooth}_{L_1}(r-\tilde{r}),
\end{split}
\end{equation}
where $L_h$ is the loss for horizontal box regression, which is the
same as that in~\cite{ren2017faster}, and ${\lambda}_{1}$,
${\lambda}_{2}$, and ${\lambda}_{3}$ are hyper-parameters that balance
the importance of each loss term.

\subsection{Inference}
\label{subsec:inference}

During testing phase, for a given image, the forward pass generates a
set of $(x, y, w, h, \alpha_1, \alpha_2, \alpha_3, \alpha_4, r)$
representing horizontal bounding boxes, four length ratios, and
obliquity factors. For each candidate, if its obliquity factor $r$ is
larger than a threshold $t_r$, indicating that the underlying object
is nearly horizontal, we select the horizontal bounding box $(x, y, w,
h)$ as the final detection. Otherwise, we select the oriented one
given by $(x, y, w, h, \alpha_1, \alpha_2, \alpha_3,
\alpha_4)$. \textcolor{black}{The non-maximum suppression (NMS) process
  is also performed. Specifically, we first adopt the efficient
  horizontal NMS (with 0.5 IoU threshold) to get rid of some candidate
  proposals, followed by an oriented NMS (with 0.1 IoU threshold) on
  the significantly reduced number of candidate proposals.}

\section{Experiments}
\label{sec:experiment}


\begin{figure*}
  \centering
  \includegraphics[width=\linewidth]{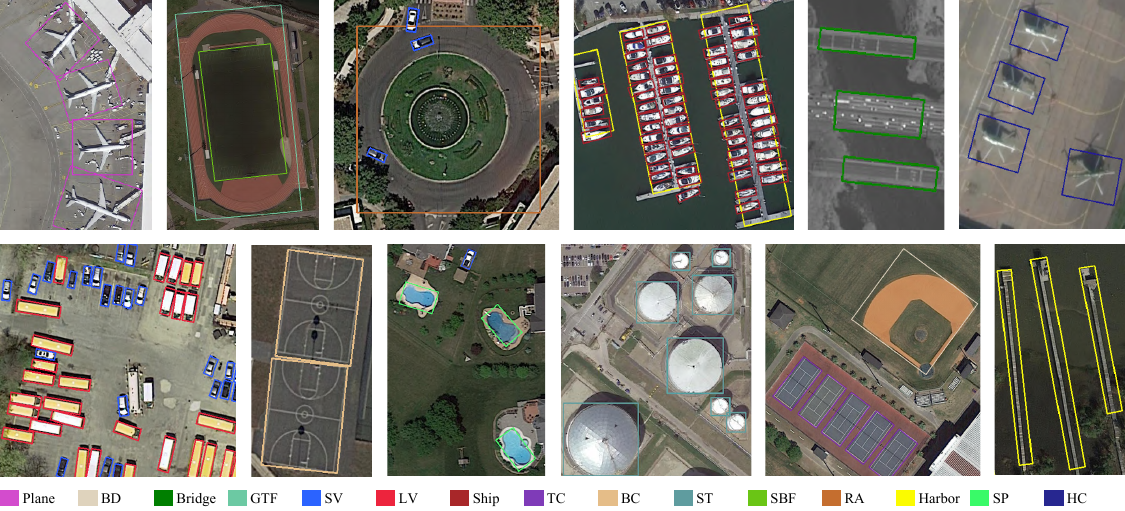}
  \vspace{-8mm}
  \caption{Some detection results of the proposed method on
    DOTA~\cite{xia2018dota}. The arbitrary-oriented objects are
    correctly detected.}
  \label{fig:overall_results}
\end{figure*}

\subsection{Datasets and evaluation protocols}

\noindent\textbf{DOTA}~\cite{xia2018dota} is a large-scale and
challenging dataset for object detection in aerial images with
quadrangle annotations. It contains 2806 $4000 \times 4000$ images and
188, 282 instances of 15 object categories: plane, baseball diamond
(BD), bridge, ground field track (GTF), small vehicle (SV), large
vehicle (LV), ship, tennis court (TC), basketball court (BC), storage
tank (ST), soccer-ball field (SBF), roundabout (RA), harbor, swimming
pool (SP) and helicopter (HC).  The official evaluation protocol of
DOTA in terms of mAP is used.

\smallskip

\noindent\textbf{HRSC2016}~\cite{liu2016ship} is dedicated for ship
detection in aerial images, containing 1061 images annotated with
rotated rectangles. We conduct experiments for the level-1 task which
detects ship from backgrounds. The standard evaluation protocol of
HRSC2016 in terms of mAP is used.

\smallskip

\noindent\textbf{MSRA-TD500}~\cite{yao2012detecting} is proposed for
detecting long and oriented texts. It contains 300 training and 200
test images annotated in terms of text lines. Since the training
set is rather small, following other methods, we also use
HUST-TR400~\cite{yao2014unified} during training. The standard
evaluation protocol of MSRA-TD500 based on F-measure is used.

\smallskip

\noindent\textbf{RCTW-17}~\cite{shi2017icdar2017} is also a long text
detection dataset, consisting of 8034 training images and 4229 test
images annotated with text lines. This dataset is very challenging due
to very large text scale variances.  We evaluate the proposed method
via the online evaluation platform in terms of F-measure.

\smallskip

\noindent\textbf{MW-18Mar}~\cite{mw18mar} is a multi-target horizontal
pedestrian tracking dataset, in which images are taken with fisheye
cameras.  \textcolor{black}{ The authors of
  \cite{tamura2019omnidirectional} extracted some frames and annotated
  the pedestrians with rotated rectangles for omnidirectional
  pedestrian detection.} The standard miss rates at
every false positive per image (FPPI) and log average miss rates
(LAMRs)~\cite{dollar2011pedestrian} are adopted for benchmarking.

\subsection{Implementation Details}

The proposed method is implemented based on the project
of
``maskrcnn\_benchmark''~\footnote{https://github.com/facebookresearch/maskrcnn-benchmark}
using 3 Titan Xp GPUs. For a fair comparison with other methods, we
adopt ResNet101~\cite{he2016resnet} for object detection in aerial
images, where the batch size is set to 6 due to limited GPU memory.
For the other experiments, ResNet50 is adopted, and the batch size is
set to 12. In all experiments, the network is trained by SGD optimizer
with momentum and weight decay set to 0.9 and $5\times10^{-4}$,
respectively. The learning rate is initialized with $7.5\times10^{-3}$
and divided by 10 at each learning rate decay step. The
hyper-parameters $\lambda_1$, $\lambda_2$, and $\lambda_3$ in
Eq.~\eqref{eq:lossreg} are set to $1$, $1$, and $16$,
respectively. Without explicitly specifying, the hyper-parameter $t_r$
on obliquity factor guiding the selection of horizontal or oriented
detection is set to 0.8. Some other application
related settings are depicted in the corresponding sections.

\textcolor{black}{We compare the proposed method with two baseline
  methods using rotated bounding box representation (denoted by RBox
  Reg.) and quadrangle representation (denoted by Vertex Reg.). For
  the RBox reg., based on horizontal prior boxes, similar
  with~\cite{ding2018transformer, liu2017rrpnship, ma2018arbitrary},
  we regress the object center $(x, y)$, long and short side length
  $(w', h')$, and the angle $\theta$ between the long side and
  X-axis. For Vertex Reg., we follow~\cite{liao2018textboxes++} by
  regressing the one-to-one vertex offset between each vertex of the
  prior box and its corresponding ground-truth vertex, which is
  ordered by minimizing the sum of vertex-wise Euclidean distances
  between the ground-truth oriented object and its horizontal bounding
  box.  For a fair comparison, both baseline methods are implemented
  using similar settings with the proposed method.}

\begin{table*}[!tbp]
  \centering
  \scriptsize
  \caption{Quantitative comparison with other methods on
    DOTA. Ours-$r$ means that the divide and conquer detection scheme
    based on obliquity factor $r$ is not used. $^*$ indicates that the
    backbone network is light-head R-CNN~\cite{li2017light}. $\dag$
    stands for evaluation using IoU threshold
    0.7. \textcolor{black}{Note that the runtime for oriented NMS is
      not included for all methods on this dataset. Otherwise, the
      proposed method using FPN runs at 9.4 FPS instead of 10.0 FPS.}
  }
  \vspace{\fixedvskip}
  \setlength{\tabcolsep}{1.1mm}{
    \label{tab:dota_quantitativeresults}
  \begin{tabular}{|c|c|cccccccccccccccc|c|}
    \hline
    Methods                                     &FPN        &Plane &BD    &Bridge &GTF   &SV    &LV    &Ship  &TC    &BC    &ST    &SBF   &RA    &Harbor &SP    &HC    &mAP   &FPS\\
    \hline\hline
    FR-O~\cite{xia2018dota}                     &-          &79.42 &77.13 &17.70  &64.05 &35.30 &38.02 &37.16 &89.41 &69.64 &59.28 &50.30 &52.91 &47.89  &47.40 &46.30 &54.13 &- \\
    RoI Trans.$^*$~\cite{ding2018transformer}   &-          &88.53 &77.91 &37.63  &74.08 &66.53 &62.97 &66.57 &90.50 &79.46 &76.75 &\textbf{59.04} &56.73 &62.54  &61.29 &55.56 &67.74 &5.9 \\
    Ours$^*$                                    &-          &\textbf{89.95} &\textbf{86.37} &45.79  &73.44 &\textbf{71.44} &68.20 &75.96 &\textbf{90.72} &\textbf{79.63} &\textbf{85.03} &58.56 &\textbf{70.19} &68.28  &\textbf{71.34} &54.45 &72.49 &8.4 \\
    Ours-$r$                                    &-          &89.93 &85.78 &45.90  &73.66 &70.07 &69.10 &76.78 &90.62 &79.08 &83.94 &57.75 &67.57 &67.53  &70.85 &56.46 &72.33 &\textbf{9.8} \\
    Ours                                        &-          &89.89 &85.99 &\textbf{46.09}  &\textbf{78.48} &70.32 &\textbf{69.44} &\textbf{76.93} &90.71 &79.36 &83.80 &57.79 &68.35 &\textbf{72.90}  &71.03 &\textbf{59.78} &\textbf{73.39} &\textbf{9.8} \\
    \hline\hline
    Azimi \etal~\cite{azimi2018towards}         &\checkmark &81.36 &74.30 &47.70  &70.32 &64.89 &67.82 &69.98 &90.76 &79.06 &78.20 &53.64 &62.90 &67.02  &64.17 &50.23 &68.16 &- \\
    RoI Trans.$^*$~\cite{ding2018transformer}   &\checkmark &88.64 &78.52 &43.44  &75.92 &68.81 &\textbf{73.68} &83.59 &90.74 &77.27 &81.46 &58.39 &53.54 &62.83  &58.93 &47.67 &69.56 &- \\
    CADNet~\cite{zhang2019cad}                  &\checkmark &87.80 &82.40 &49.40  &73.50 &71.10 &63.50 &76.60 &\textbf{90.90} &79.20 &73.30 &48.40 &60.90 &62.00  &67.00 &\textbf{62.20} &69.90 &- \\
    R$^2$CNN++~\cite{yang2018r2cnn++}           &\checkmark &89.66 &81.22 &45.50  &75.10 &68.27 &60.17 &66.83 &\textbf{90.90} &80.69 &86.15 &\textbf{64.05} &63.48 &65.34  &68.01 &62.05 &71.16 &- \\
    RBox reg.                                   &\checkmark &89.37 &75.96 &35.43 &69.57 &68.35 &63.78 &74.92 &90.76 &\textbf{84.70} &85.26 &62.43 &62.40 &52.97  &60.32 &54.61 &68.72 &9.2 \\
    Vertex reg.                                 &\checkmark &80.16 &76.77 &43.31 &69.38 &55.71 &56.52 &72.25 &88.10 &28.95 &86.31 &63.66 &62.23 &61.62  &68.18 &41.65 &63.65 &9.8 \\
    Ours$^*$                                    &\checkmark &\textbf{90.02} &84.41 &49.80 &\textbf{77.93} &72.23 &72.52 &85.81 &90.85 &79.21 &86.61 &59.01 &69.15 &66.30  &71.22 &55.67 &74.05 &7.1 \\
    Ours-$r$                                    &\checkmark &89.40 &\textbf{85.08} &52.00 &77.40 &72.68 &72.89 &86.41 &90.74 &78.80 &86.79 &57.84 &70.42 &67.73  &\textbf{71.64} &56.63 &74.43 &\textbf{10.0} \\
    Ours                                        &\checkmark &89.64 &85.00 &\textbf{52.26} &77.34 &\textbf{73.01} &73.14 &\textbf{86.82} &90.74 &79.02 &\textbf{86.81} &59.55 &\textbf{70.91} &\textbf{72.94} &70.86 &57.32 &\textbf{75.02} &\textbf{10.0} \\
    \hline\hline
    RBox reg.$\dag$                             &\checkmark &42.52 &21.76 &10.47 &36.53 &26.57 &26.91 &32.39 &63.20 &36.56 &33.54 &33.04 &15.63  &11.16 &10.05 &12.98 &27.56 &9.2 \\
    Vertex reg.$\dag$                           &\checkmark &67.94 &50.51 &14.28 &47.46 &29.79 &27.92 &40.66 &72.75 &14.29 &67.59 &33.47 &40.87  &22.04 &17.91 &15.13 &37.51 &9.8 \\
    Ours$^*\dag$                                &\checkmark &\textbf{77.98} &53.21 &12.52 &\textbf{68.87} &47.25 &46.07 &54.83 &\textbf{90.45} &68.00 &68.45 &\textbf{56.44} &40.12 &28.59  &22.47 &19.13 &50.29 &7.1 \\
    Ours-$r\dag$                                &\checkmark &67.66 &50.37 &\textbf{17.07} &60.60 &48.74 &49.00 &61.59 &88.98 &68.84 &\textbf{74.83} &48.30 &48.03 &32.58  &23.78 &23.75 &50.94 &\textbf{10.0} \\
    Ours$\dag$                                  &\checkmark &77.32 &\textbf{59.75} &15.95  &67.63 &\textbf{50.02} &\textbf{50.25} &\textbf{63.62} &90.38 &\textbf{69.04} &74.56 &51.58 &\textbf{50.16} &\textbf{32.73} &\textbf{24.19} &\textbf{25.18} &\textbf{53.49} &\textbf{10.0} \\
    \hline
  \end{tabular}
  \vspace{\fixedvskip}
  }
\end{table*}

\begin{table}[!tbp]
  \centering
  \scriptsize
    \caption{Quantitative comparison with some state-of-the-art
      methods on HRSC2016. $^*$ indicates that Light-head R-CNN is
      adopted.}
    \vspace{\fixedvskip}
    \setlength{\tabcolsep}{1.1mm}{
    \label{tab:hrsc_quantitativeresults}
    \begin{tabular}{|c|cccc|cc|}
    \hline
    Methods & RC2~\cite{liu2017rotated} &R$^2$PN~\cite{zhang2018toward} &RRD~\cite{liao2018rotation} &RoI Trans.$^*$~\cite{ding2018transformer} &Ours$^*$  &Ours  \\ \hline
      mAP      &75.7 &79.6 &84.3 &86.2 &87.4 &\textbf{88.2}   \\ \hline
    \end{tabular}
    \vspace{\fixedvskip}
    }
\end{table}

\subsection{Object detection in aerial images}
For the experiments on DOTA~\cite{xia2018dota}, we train the model for
$50k$ steps, and the learning rate decays at $\{38k, 46k\}$
steps. Random rotation with angle among $\{0, \pi/2, \pi, 3\pi/2\}$
and class balancing are adopted for data augmentation. For the
experiments on HRSC2016~\cite{liu2016ship}, we train the model for
$3.2k$ steps and decay the learning rate at $2.8k$ steps. Horizontal
flipping is applied for data augmentation. For a fair comparison, the
size of training/test images and the anchor settings on both datasets
are kept the same as~\cite{ding2018transformer}.

\begin{figure*}
  \centering
  \subfloat[]{
    \label{fig:result_hrsc1}
    \includegraphics[height=4cm,keepaspectratio]{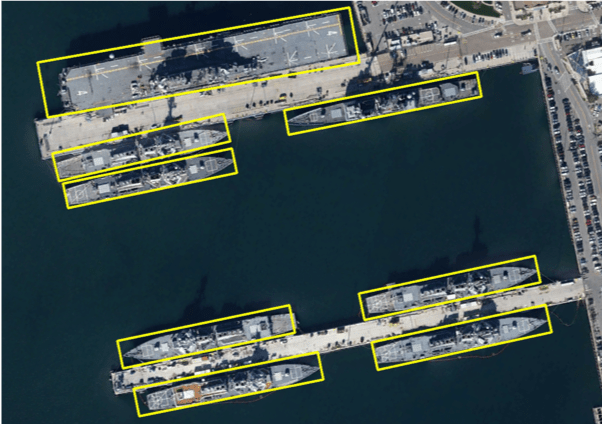}
  }
  \subfloat[]{
    \label{fig:result_td1}
    \includegraphics[height=4cm,keepaspectratio]{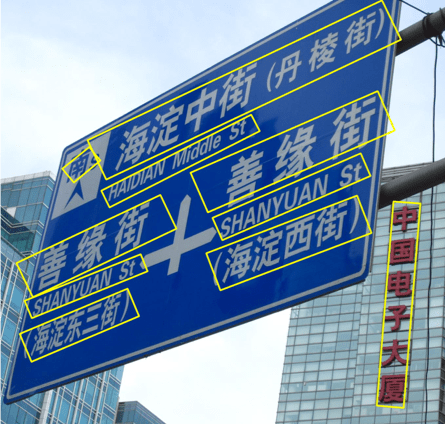}
  }
  \subfloat[]{
    \label{fig:result_td2}
    \includegraphics[height=4cm,keepaspectratio]{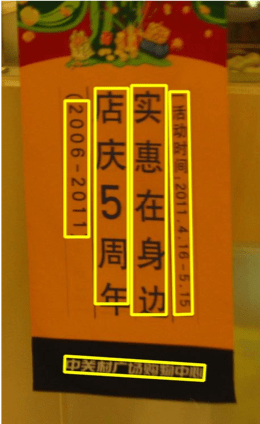}
  }
  \subfloat[]{
    \label{fig:result_rctw1}
    \includegraphics[height=4cm,keepaspectratio]{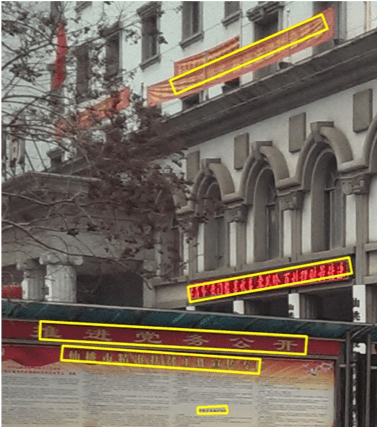}
  }
  \subfloat[]{
    \label{fig:result_rctw2}
    \includegraphics[height=4cm,keepaspectratio]{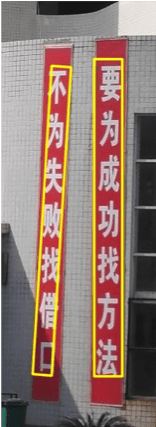}
  }
  \vspace{-2mm}
  \caption{Some detection results of the proposed method on
    HRSC2016~\cite{liu2016ship} in (a),
    MSRA-TD500~\cite{yao2012detecting} in (b-c), and
    RCTW-17~\cite{shi2017icdar2017} in (d-e).
  }
  \label{fig:vis_text&hrsc}
\end{figure*}

\smallskip
\noindent\textbf{Overall results}. Some qualitative results on DOTA
and HRSC2016 are shown in Fig.~\ref{fig:overall_results} and
Fig.~\ref{fig:vis_text&hrsc}(a), respectively. We show all detected
objects with classification scores above 0.6. As illustrated, the
proposed method accurately detects both horizontal and oriented
objects even under dense distribution and/or being long. The
quantitative comparisons with other methods on DOTA~\cite{xia2018dota}
and HRSC2016~\cite{liu2016ship} are depicted in
Tab.~\ref{tab:dota_quantitativeresults} and
Tab.~\ref{tab:hrsc_quantitativeresults}, respectively. Without any
extra network design such as cascade refinement and attention
mechanism, the proposed method outperforms some state-of-the-art
methods on both DOTA and HRSC2016 and is more efficient in
runtime. Specifically, For the experiment on DOTA, the proposed method
without FPN~\cite{lin2017feature} achieves 73.39\% mAP, outperforming
the state-of-the-art method~\cite{ding2018transformer} by 5.65\%
mAP. FPN~\cite{lin2017feature} that exploits better multi-scale
features is also beneficial for the proposed method, boosting the
performance to 75.02\%. The proposed method using
FPN~\cite{lin2017feature} improves the state-of-the-art
method~\cite{yang2018r2cnn++} by 3.86\% mAP. For HRSC2016 dataset, the
proposed method achieves 88.2\% mAP, improving state-of-the-art
methods by 2\%.

\smallskip
\noindent\textbf{Experiments on different network architectures}. To
further demonstrate the versatility of the proposed method, we
evaluate the proposed method on different networks. Concretely, we
replace the faster R-CNN head by light-head R-CNN~\cite{li2017light}
head. As depicted in Tab.~\ref{tab:dota_quantitativeresults}, using
the same network on DOTA~\cite{xia2018dota}, the proposed method
improves~\cite{ding2018transformer} by 4.49\% and 4.75\% mAP with and
without FPN, respectively. The proposed method
outperforms~\cite{ding2018transformer} by 1.2\% mAP on
HRSC2016~\cite{liu2016ship}.

\begin{figure}[t]
  \centering
  \hspace*{-0.46em}
  \includegraphics[width=0.95\linewidth]{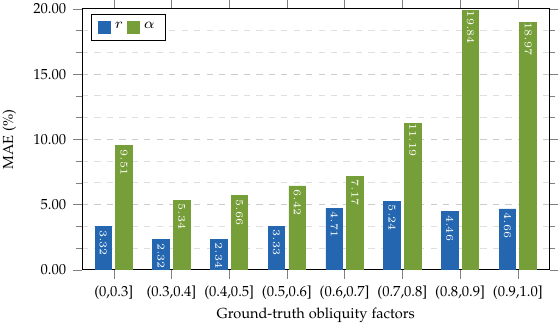}
  \vspace{\fixedvskip}
  \caption{\textcolor{black}{Mean absolute error (MAE) of obliquity factor $r$ and gliding offset $\alpha$ regression with respect to different ranges of ground-truth obliquity factors for the proposed method on DOTA.}}
\label{fig:alpha_r}
\end{figure}

\smallskip

\noindent\textbf{Ablation study}. \textcolor{black}{We conduct ablation
  study on DOTA~\cite{xia2018dota}. The proposed method relies on a
  novel multi-oriented object representation composed of three
  components: horizontal bounding box $(x, y, w, h)$, gliding offsets
  $(\alpha_1, \alpha_2, \alpha_3, \alpha_4)$, and obliquity factor
  $r$. We begin with analyzing the quality of each individual
  component using Faster R-CNN head with FPN. Firstly, the proposed
  method achieves a good performance with 76.22\% mAP under horizontal
  bounding box evaluation. The small performance gap (\ie, 1.2\% mAP)
  between oriented and horizontal object detection implies that the
  gliding offset regression is also quite accurate. We also explicitly
  evaluate the accuracy of gliding offset regression in terms of mean
  absolute error (MAE) for the correctly detected objects. As depicted
  in Fig.~\ref{fig:alpha_r}, the gliding offset regression is quite
  accurate for oriented objects, but is less precise for nearly
  horizontal objects (\eg, $\tilde{r} > 0.8$) for which potential
  confusion issue remains. This motivates us to regress the obliquity
  factor $r$ to guide the selection of horizontal or oriented
  detection as the final detection result, helping to remedy the
  remaining confusion issue for nearly horizontal objects. Indeed, as
  shown in Fig.~\ref{fig:alpha_r}, the obliquity factor $r$ regression
  is in general very accurate (MAE $< 5.3\%$). This quality analysis of
  each individual component of the proposed multi-oriented object
  representation confirms the effectiveness of the proposed method.}

\begin{figure}
 \setlength\tabcolsep{1pt}
 \settowidth\rotheadsize{vertices}
 \hspace*{-5pt}
 \begin{tabularx}{\linewidth}{c cccc }
   \rothead{RBox reg.}         & \includegraphics[width=2cm, height=1.9cm, valign=m]{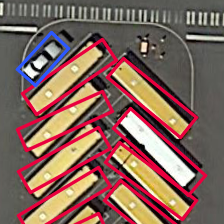}
                               & \includegraphics[width=2cm, height=1.9cm, valign=m]{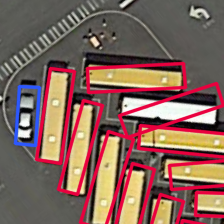}
                               & \includegraphics[width=2cm, height=1.9cm, valign=m]{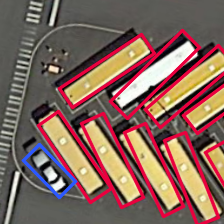}
                               & \includegraphics[width=2cm, height=1.9cm, valign=m]{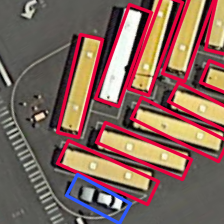} \\ \addlinespace[2pt]
   \rothead{Vertex reg.}     & \includegraphics[width=2cm, height=1.9cm, valign=m]{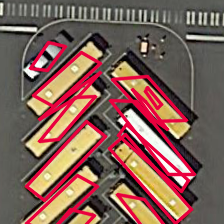}
                               & \includegraphics[width=2cm, height=1.9cm, valign=m]{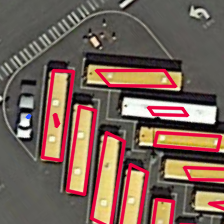}
                               & \includegraphics[width=2cm, height=1.9cm, valign=m]{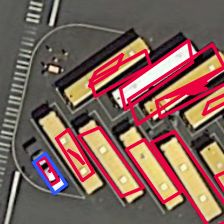}
                               & \includegraphics[width=2cm, height=1.9cm, valign=m]{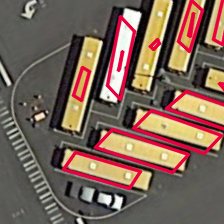} \\ \addlinespace[2pt]
   \rothead{Ours}              & \includegraphics[width=2cm, height=1.9cm, valign=m]{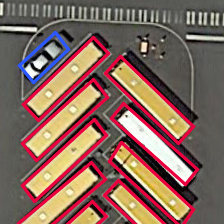}
                               & \includegraphics[width=2cm, height=1.9cm, valign=m]{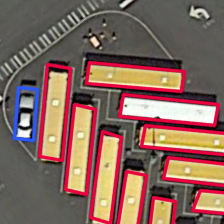}
                               & \includegraphics[width=2cm, height=1.9cm, valign=m]{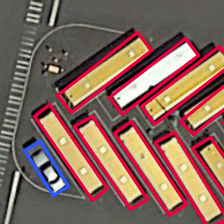}
                               & \includegraphics[width=2cm, height=1.9cm, valign=m]{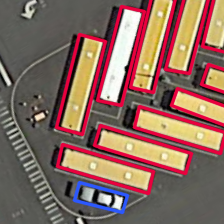} \\ \addlinespace[2pt]
   \rothead{}                  & 0\degree
                               & 40\degree
                               & 80\degree
                               & 120\degree
 \end{tabularx}
 \vspace{\fixedvskip}
 \caption{Qualitative comparison with baseline methods in detecting
   objects of different orientations (by rotating an input image with
   different angles). The meaning of colors is the same as that in
   Fig.~\ref{fig:overall_results}.}
 \label{fig:ablation}
\end{figure}

Some qualitative comparison can be found in
Fig.~\ref{fig:ablation}. We rotate an image with several different
angles and test the proposed method and two baseline methods on the
rotated images. \textcolor{black}{The RBox reg.}  produces inaccurate
results due to the imprecise angle regression. \textcolor{black}{The Vertex
  reg.} have difficulty for tilted objects at some orientations due to
the confusion in defining the vertex order in training. The proposed
method is able to accurately detect objects of any orientations.

The quantitative comparison with baseline methods is depicted in the
middle of Tab.~\ref{tab:dota_quantitativeresults}. The proposed method
outperforms the two baseline methods by a large margin. Specifically,
the proposed method outperforms the \textcolor{black}{RBox reg. and
  Vertex reg.} by 6.30\% and 11.37\% mAP at the cost of ignorable
runtime. In fact, as depicted in
Tab.~\ref{tab:dota_quantitativeresults}, the proposed method is more
efficient than both baseline methods producing more false
detections. To further demonstrate the accuracy of the proposed
method, we also conduct a benchmark using larger IoU threshold 0.7 in
the evaluation system. As shown in
Tab.~\ref{tab:dota_quantitativeresults}, the improvement is even more
significant, changing from 6.30\% (\resp 11.37\%) to 25.93\% (\resp
15.98\%). \textcolor{black}{This further demonstrates the accuracy of
  the proposed method in detecting oriented objects.}


We then assess the individual contribution of the proposed
\textcolor{black}{vertex gliding} and divide-and-conquer detection
scheme in the proposed method for \textcolor{black}{multi-oriented
  object detection}. To this end, we evaluate an alternative of the
proposed method by discarding the divide-and-conquer detection scheme
based on obliquity factor $r$. As depicted in
Tab.~\ref{tab:dota_quantitativeresults}, the proposed representation
in terms of $(x, y, w, h, \alpha_1, \alpha_2, \alpha_3, \alpha_4)$
contributes a lot to the improvement. The proposed detection scheme
brings 0.59\% and 1.06\% mAP improvement with and without
FPN~\cite{lin2017feature}, respectively. When larger IoU threshold 0.7
is used, the selection scheme yields 2.55\% mAP improvement,
\textcolor{black}{confirming the effectiveness of the selection scheme
  based on obliquity factor $r$. Without the selection scheme, some
  nearly horizontal objects with inaccurate predicted gliding offsets
  (see Fig.~\ref{fig:alpha_r}) may be considered as correct
  (\textit{resp.} incorrect) detection under evaluation with 0.5
  (\textit{resp.} 0.7) IoU threshold. This explains the more
  significant improvement of the selection scheme when a larger IoU
  threshold is used for evaluation.}

\textcolor{black} {We also analyze the effect of different thresholds
  $t_r$ of obliquity factor $r$ on DOTA dataset using Faster R-CNN
  head with FPN.  As depicted in Tab.~\ref{tab:ablation_r}, the
  performance is rather stable, especially for $t_r \in [0.75,
    0.85]$. The performance slightly decreases for smaller and larger
  $t_r$. Indeed, with a very small threshold $t_r$, horizontal bounding
  boxes are selected to represent some oriented objects, which leads
  to inaccurate detection. When a large threshold $t_r$ is adopted,
  the potential confusion issue for nearly horizontal objects remains,
  also resulting in decreased performance.}

\begin{table}[!tbp]
  \centering
  \scriptsize
  \caption{\textcolor{black}{Ablation study on different thresholds $t_r$ of obliquity factor $r$.}}
  \vspace{\fixedvskip}
    \setlength{\tabcolsep}{1.1mm}{
    \label{tab:ablation_r}
    \begin{tabular}{|c|ccccccc|}
    \hline
      $t_r$         &0.65  &0.70  &0.75  &0.80  &0.85  &0.90 &0.95  \\ \hline
      w FPN    &73.29 &74.30 &74.72 &75.02 &\textbf{75.06} &\textbf{75.06} &74.44   \\ \hline
      w/o FPN  &71.76 &72.42 &73.24 &\textbf{73.39} &73.37 &72.59 &72.47   \\ \hline
    \end{tabular}
    \vspace{\fixedvskip}
    }
\end{table}



\subsection{Long text detection in natural scenes}
For oriented scene text detection on
MSRA-TD500~\cite{yao2012detecting} and
RCTW-17~\cite{shi2017icdar2017}, we apply the same data augmentation
as SSD~\cite{liu2016ssd}. Besides, we also randomly rotate the images
with $\pi/2$ to better handle vertical texts. The training images are
randomly cropped and resized to some specific sizes. For MSRA-TD500,
we randomly resize the short side of cropped images to
$\{512, 768, 864\}$. For RCTW-17~\cite{shi2017icdar2017} containing
many small texts, the short side is
randomly resized to $\{960, 1200, 1400\}$. We first pre-train the
model on SynthText~\cite{gupta2016synthetic} for one epoch. Then we
fine-tune the model for $4k$ (\resp $14k$) and decay the learning rate
at $3k$ (\resp $10k$) steps for MSRA-TD500 (\resp RCTW-17). During
test, the short side of MSRA-TD500 images is resized to 768. For
RCTW-17, the short side is set to 1200 for single scale test. We add
extra scales of $\{512, 1024, 1280, 1560\}$ for multi-scale test.

Some qualitative illustrations are given in
Fig.~\ref{fig:vis_text&hrsc}(b-e). The proposed method correctly
detect texts of arbitrary orientations. The quantitative comparisons
with some state-of-the-art methods on MSRA-TD500 and RCTW-17 are
depicted in Tab.~\ref{tab:td500_quantitativeresults} and
Tab.~\ref{tab:rctw_quantitativeresults}, respectively.
The proposed method outperforms other competing methods and is more
efficient on both datasets. Specifically, on MSRA-TD500,
the proposed method under single scale test outperforms the
multi-scale version of~\cite{he2018direct} using larger extra training
images by 0.5\%, and improves~\cite{wang2019arbitrary} by 2.9\%. On
RCTW-17, the proposed method outperforms the state-of-the-art
method~\cite{zhang2019look} by 5.8\% (\resp 0.9\%) under single-scale
(\resp multi-scale) test while being much more efficient.

\begin{table}[!tbp]
  \centering
  \caption{Quantitative comparison with other methods on
    MSRA-TD500~\cite{yao2012detecting}. MS stands for multi-scale
    test.
  }
  \vspace{\fixedvskip}
  \setlength{\tabcolsep}{1.1mm}{
    \label{tab:td500_quantitativeresults}
  \begin{tabular}{|c|c|c|c|c|}
    \hline
    Methods                       & Precision   & Recall    & F-measure & FPS  \\
    \hline
    Zhang \etal~\cite{zhang2016multi}                  & 83.0        & 67.0      & 74.0 & 0.5 \\
    SegLink~\cite{shi2017seglink}                       & 86.0        & 70.0      & 77.0 & 8.9 \\
    RRD~\cite{liao2018rotation}                           & 87.0        & 73.0      & 79.0 & 10.0 \\
    EAST~\cite{zhou2017east}                         & 87.3       & 67.4     & 76.1 & 13.2 \\
    Border MS~\cite{xue2018accurate}                        & 83.0        & 73.3      & 76.8 & - \\
    TextField~\cite{xu2019textfield}                    & 87.4        & 75.9      & 81.3  & 5.2 \\
    Lyu \etal~\cite{lyu2018multi}                    & 87.6        & 76.2      & 81.5 & 5.7 \\
    \textcolor{black}{CRAFT~\cite{baek2019character}}                   & \textcolor{black}{88.2}       & \textcolor{black}{78.2}       & \textcolor{black}{82.9} & \textcolor{black}{8.6} \\
    MCN~\cite{liu2018mcn}                           & 88.0        & 79.0      & 83.0 & - \\
    Wang \etal~\cite{wang2019arbitrary}                           & 85.2        & 82.1      & 83.6 & 10.0 \\
    Direct MS~\cite{he2018direct}               & \textbf{91.0}    & 81.0 &86.0  &- \\
    Ours                          & 88.8    & \textbf{84.3}   &\textbf{86.5}  &\textbf{15.0} \\
    \hline
  \end{tabular}
  \vspace{\fixedvskip}
  }
\end{table}

\begin{table}[!tbp]
  \centering
  \caption{Quantitative comparison with other methods on
    RCTW-17~\cite{shi2017icdar2017}. MS stands for multi-scale
    test. 
  }
  \vspace{\fixedvskip}
  \setlength{\tabcolsep}{1.1mm}{
    \label{tab:rctw_quantitativeresults}
  \begin{tabular}{|c|c|c|c|c|}
    \hline
    Methods                       & Precision   & Recall    & F-measure & FPS  \\
    \hline
    Official baseline~\cite{shi2017icdar2017}                  & 76.0        & 40.4      & 52.8  & 8.9 \\
          RRD~\cite{liao2018rotation}                           & 72.4       & 45.3      & 55.7 & 10.0 \\
    RRD MS                           & 77.5        & 59.1      & 67.0 & - \\
    Direct MS~\cite{he2018direct}                          & 76.7       & 57.9     & 66.0 & - \\
    Border MS~\cite{xue2018accurate}                       & 78.2        & 58.8      & 67.1 & - \\
          LOMO~\cite{zhang2019look}             & \textbf{80.4}      & 50.8          & 62.3 &4.4 \\
    LOMO MS                         & 79.1      & 60.2          & 68.4 &- \\
    Ours                          & 77.0    &61.0          &68.1  &7.8 \\
    Ours MS                         & 77.6    &\textbf{62.7} &\textbf{69.3}  &- \\
    \hline
  \end{tabular}
  \vspace{\fixedvskip}
  }
\end{table}

\subsection{Pedestrian detection in fisheye images}
\label{subsec:pedestrian_det_exp}

We compare the proposed method with \textcolor{black}{the two} baseline
methods RBox reg. and Vertex reg., classical horizontal box regression
\textcolor{black}{(denoted by HBox reg.), and the method in
  \cite{tamura2019omnidirectional}} on
MW-18Mar~\cite{mw18mar}. \textcolor{black}{For a fair comparison
  with~\cite{tamura2019omnidirectional}, we follow similar training
  and test settings
  with~\cite{tamura2019omnidirectional}. Specifically, in all
  experiments, FPN is not used. All images are resized to $416 \times
  416$ during training and test. During training,} We randomly rotate
the images for data augmentation. The model is trained in total for
$4k$ steps and the learning rate decays at $3k$ steps.

\begin{figure}[t]
  \centering
  \hspace*{-0.51em}
  \subfloat[HBox reg.]{
    \label{fig:mw2_hbox}
    \includegraphics[height=3.1cm,keepaspectratio]{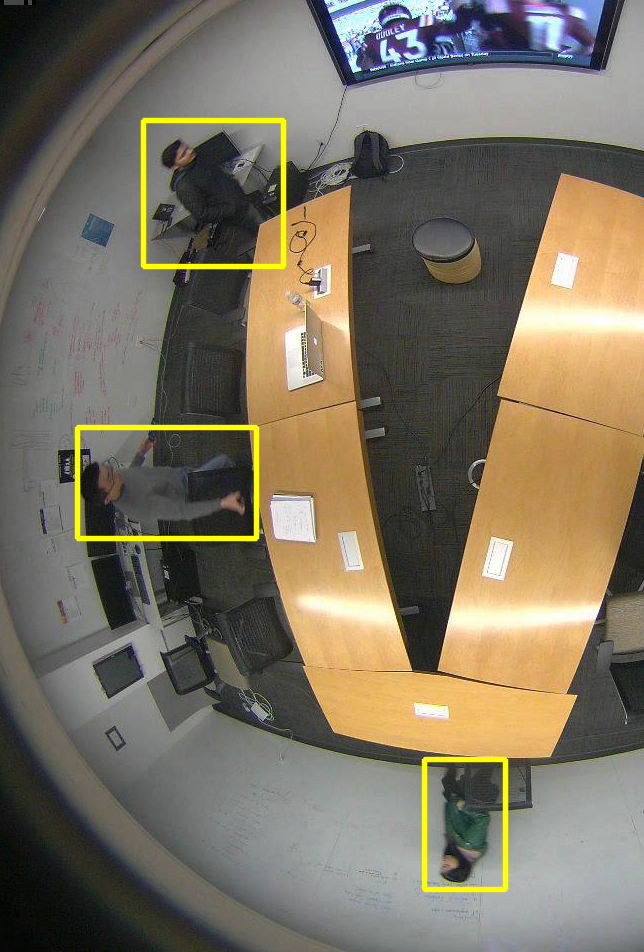}
  }
  \hspace*{-0.46em}
  \subfloat[RBox reg.]{
    \label{fig:mw2_rbox}
    \includegraphics[height=3.1cm,keepaspectratio]{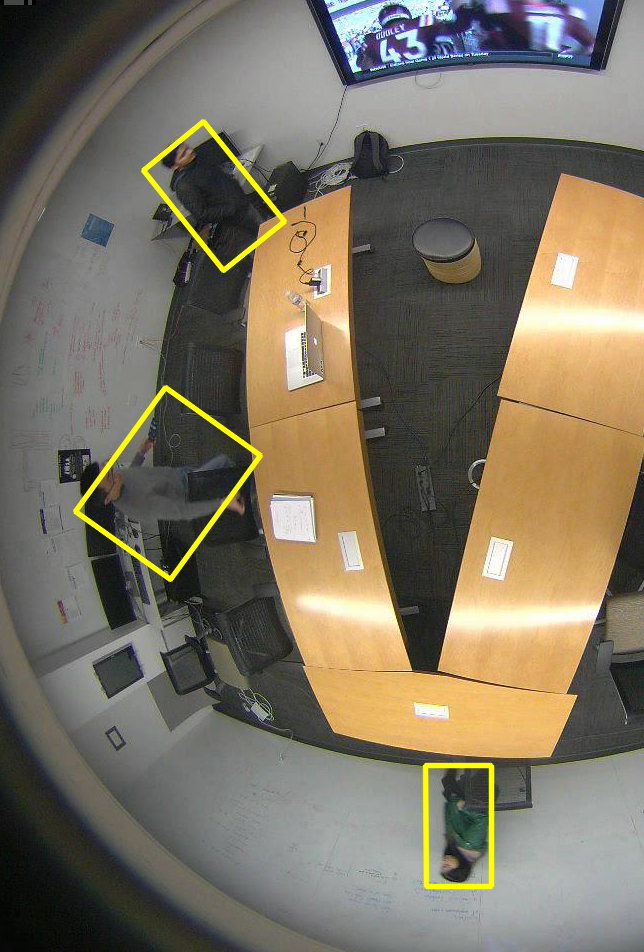}
  }
  \hspace*{-0.46em}
  \subfloat[Vertex reg.]{
    \label{fig:mw2_quad}
    \includegraphics[height=3.1cm,keepaspectratio]{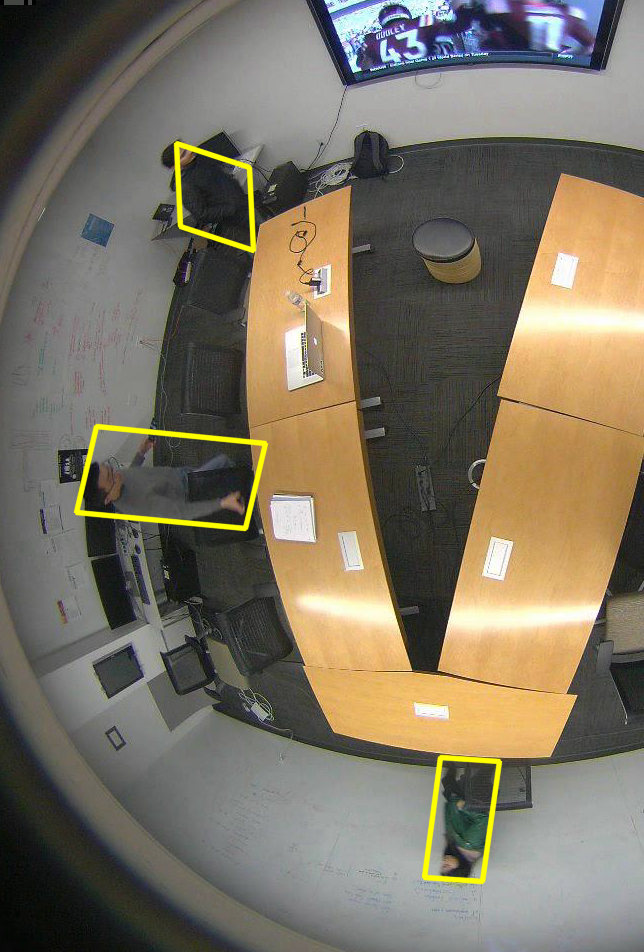}
  }
  \hspace*{-0.46em}
  \subfloat[Ours]{
    \label{fig:mw2_our}
    \includegraphics[height=3.1cm,keepaspectratio]{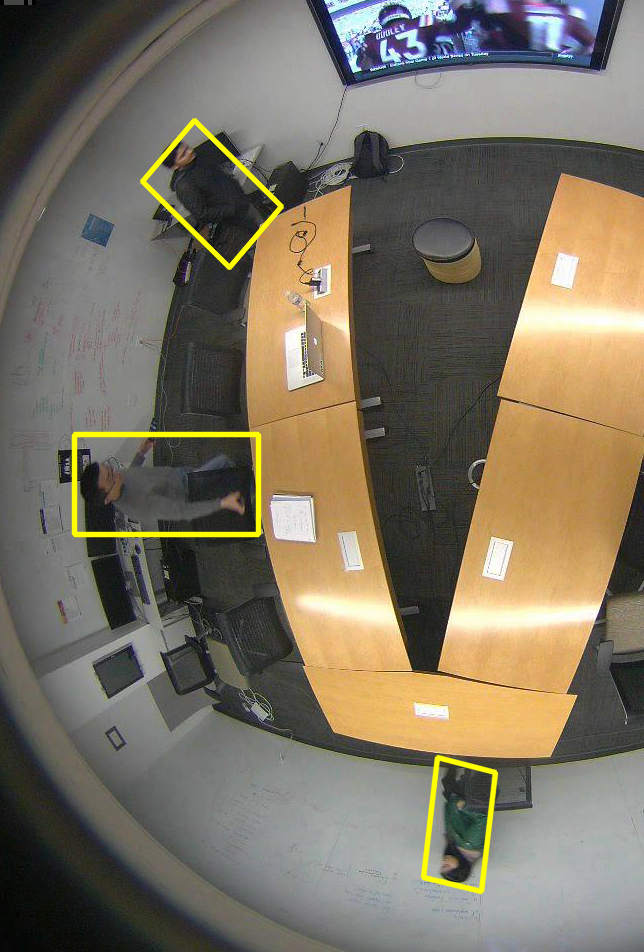}
  }
  \vspace{-2mm}
  \caption{Qualitative illustrations of different methods on MW-18Mar~\cite{mw18mar}.}
  \label{fig:vis_mw2}
\end{figure}

\begin{figure}[t]
  \centering
  \hspace*{-0.46em}
  \includegraphics[width=0.95\linewidth]{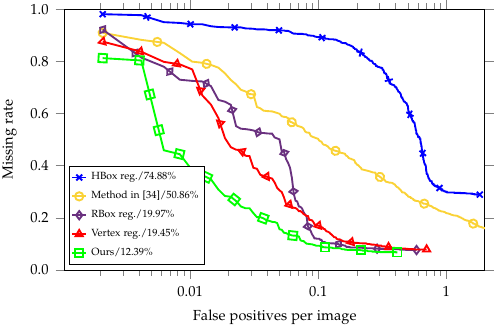}
  \vspace{\fixedvskip}
  \caption{Evaluation on MW-18Mar~\cite{mw18mar}. The numbers are the
    LAMRs.
  }
\label{fig:fppi_of_mw18mar2}
\end{figure}

Some qualitative results are illustrated in
Fig.~\ref{fig:vis_mw2}.
The proposed method achieves more accurate results than all the baseline methods.
The curve of missing rate with respect to the number of false positives per image is
depicted in Fig.~\ref{fig:fppi_of_mw18mar2}. The proposed
method achieves lower missing rate \textcolor{black}{than all the other methods}.




\section{Conclusion}
\label{sec:conclusion}
In this paper, we propose a simple yet effective representation for
oriented objects and a divide-and-conquer strategy to detect
multi-oriented objects.  Based on this, we build a robust and fast
multi-oriented object detector. It accurately detects ubiquitous
multi-oriented objects such as objects in arial images, scene texts,
and pedestrians in fisheye images.  Extensive experiments demonstrate
that the proposed method outperforms some state-of-the-art methods on
multiple benchmarks while being more efficient. In the future, we
would like to explore the complementary of the proposed method with
other approaches focusing on feature enhancement. One-stage
multi-oriented object detector is also another direction which is
worthy of exploitation.

\ifCLASSOPTIONcaptionsoff
  \newpage
\fi

\renewcommand{\baselinestretch}{.95}

\bibliographystyle{IEEEtran}

\bibliography{reference}

\end{document}